# Sampling First Order Logical Particles

**Hannaneh Hajishirzi and Eyal Amir**
Department of Computer Science
University of Illinois at Urbana-Champaign
{hajishir, eyal}@uiuc.edu

## Abstract

Approximate inference in dynamic systems is the problem of estimating the state of the system given a sequence of actions and partial observations. High precision estimation is fundamental in many applications like diagnosis, natural language processing, tracking, planning, and robotics. In this paper we present an algorithm that samples possible deterministic executions of a probabilistic sequence. The algorithm takes advantage of a compact representation (using first order logic) for actions and world states to improve the precision of its estimation. Theoretical and empirical results show that the algorithm's expected error is smaller than propositional sampling and Sequential Monte Carlo (SMC) sampling techniques.

## 1 Introduction

Important AI applications like natural language processing, program verification, tracking, and robotics involve stochastic dynamic systems. The current state of such systems is changed by executing actions. Reasoning is the task of computing the posterior probability over the state of such dynamic systems given past actions and observations. Reasoning is difficult because the system's exact initial state or the effects of its actions are uncertain (e.g., there may be some noise in the system or its actions may fail).

Exact reasoning (e.g., [11; 1]) is not tractable for long sequence of actions in complex systems. This is because domain features become correlated after some steps, even if the domain has much conditional-independence structure [4]. Therefore, approximate reasoning is of much interest. One of the most commonly used classes of techniques for approximate reasoning is SMC sampling [5]. These methods are efficient, but they require many samples (exponential in the dimensionality of the domain) to yield a lower error rate. Recently, [8] introduced a new sampling approach which achieves higher precision than SMC techniques given a fixed number of samples. Still, it requires a large number of samples in complex domains because in their method the domains are represented using propositional logic.

In this paper we present a sampling-based filtering algorithm in a first order dynamic system called *Probabilistic Relational Action Model (PRAM)*. We show that our new algorithm takes fewer samples and yields better accuracy than previous sampling techniques. Such improvement is possible because of the underlying deterministic structure of the transition system, the compact representation of the domains using first order logic (FOL), and efficient subroutines for first order logical *regression* (e.g., [17]) and first order logical *filtering* [19; 14].

We model a PRAM (Section 2) using probabilistic situation calculus [17], extended with a first order probabilistic prior that combines FOL and probabilities in a single framework (e.g., [18; 3; 12; 15; 7; 20]). Transitions in a PRAM are modeled with probability distributions over possible deterministic executions of probabilistic actions (every transition model can be represented this way).

Our algorithm (Section 3) first samples sequences of deterministic actions, called *first order (FO) particles*, that are possible executions of the given probabilistic action sequence and are consistent with the observations. It also updates the current state of the system given each FO particle. Next, the algorithm computes the probability of the query given the updated current state. To do so, it applies logical regression to the query for each FO particle, computes a FOL formula that represents all the possible initial states, and computes the prior probability of that formula. Finally, the algorithm computes the posterior probability as the weighted sum of these derived prior probabilities.

This algorithm achieves superior precision with fewer samples than SMC sampling techniques [5]. The intuition behind this improvement is that each FO particle corresponds to exponentially many state sequences (particles) generated by earlier techniques. The algorithm is computationally efficient when FOL regression and filtering with the deterministic actions are efficient. We prove our claim about precision and verify the results empirically by several experiments (Section 4).

Our representation for PRAM differs from Dynamic Bayesian Networks (DBNs) (e.g., [13]). DBNs emphasize conditional independence among random variables. In contrast, our model applies a representation for the transition model as a distribution over deterministic actions. Also, PRAM uses a different model for observations. The observations are received asynchronously without prediction of what will be observed. Both frameworks are universal and can represent each other, but they are more compact and natural in different scenarios.

The closest work to ours is [8] which also samples deterministic executions of the given probabilistic sequence. However, that work assumes that the deterministic sequences are always executable. In this paper we relax this assumption and avoid sampling the sequences that are not executable by keeping track of the current state of the system. Also, our representation is more compact (uses FOL), so it achieves higher precision with less samples.

Recently, [10; 22] explore reasoning algorithms in relational HMMs. They use a different representation than ours for their observations. Moreover, they do not use a compact representation for their prior distribution. Earlier algorithm trades efficiency of computation for precision as it is an exact method. Latter, introduced a sampling algorithm where each sample represents a set of states and is generated from the probability distribution over disjoint sets of the states. Therefore, to update this distribution they build new disjoint sets at each time step which leads to a large number of disjoint sets (exponential in domain features) in long sequences. Our algorithm is different in a sense that we do not sample states; we sample deterministic sequences which correspond to state sequences that are derived by regressing the query through the deterministic sequence.

## 2 Probabilistic Relational Action Models

In this section we present our framework, called *Probabilistic Relational Action Model* (PRAM), for representing the dynamic system. The system is dynamic in a sense that its state changes by executing actions. Actions have probabilistic effects that are represented with a probability distribution over possible deterministic executions.

A PRAM consists of two main parts: (1) A prior knowledge representing a probability distribution $P^0$ over initial world states in a relational model. (2) A probability distribution *PA* over deterministic executions of probabilistic actions. In what follows we define the basic building blocks of a PRAM. We first define the language $\mathcal{L}$ of a PRAM for representing the probability distributions:

**Definition 1.** [PRAM language] The language $\mathcal{L}$ of a PRAM is a tuple $(F, C, V, \mathcal{A}, D\mathcal{A})$ consisting of:

- $F$ a finite set of predicate variables (called *fluents*) whose values change over time.
- $C$ a finite set of constants representing objects in the domain.
- $V$ a finite set of variables.
- $\mathcal{A}, D\mathcal{A}$ finite sets of probabilistic and deterministic action names, respectively.

We define a *fluent atom* as a formula of the form $f(x_1, \ldots, x_k)$ (also represented by $f(\vec{x})$), where $x_1, \ldots, x_k \in V \cup C$ are either variables or constants. In PRAM grounding of a fluent $f(\vec{x})$ is defined as replacing each variable in $\vec{x}$ with a constant $c \in C$. Accordingly, a world state $s \in \mathcal{S}$ is defined as a full assignment of $\{true, false\}$ to all the groundings of all the fluents in $F$.

**Example 1** (Briefcase). *Briefcase* is a domain consisting of *objects*, *locations*, and a *briefcase* $B$. The variables ?o and ?l denote *objects* and *locations*, respectively. The fluents are: $In(?o), At(B, ?l), At(?o, ?l)$. An agent interacts with the system by executing some probabilistic actions: *putting* objects in, *PutIn*(?o, B), *taking* objects *out* of the briefcase, *TakeOut*(?o, B), and *moving* the briefcase with objects inside, *Move*($B, ?l_1, ?l_2$). Some deterministic actions in the domain are: *MvWithObj*, *PutInSucc*, and *PutInFail*.

In PRAM each deterministic action $da(\vec{x}) \in D\mathcal{A}$ is specified by *precondition* and *successor state* axioms [17].

**Definition 2.** [Deterministic action axioms] Precondition axioms show the conditions under which the deterministic action $da$ is executable in a given state; *Precond* is a special predicate denoting the executability of a deterministic action: $Precond_{da}(\vec{x}) \Leftrightarrow \Phi(\vec{x})$ where $\Phi$ is a FOL formula.

Successor state axioms enumerate all the ways that the value of a particular fluent can be changed; A successor state axiom for a fluent $f$ is defined as:

$$Precond_{da}^t(\vec{x}) \Rightarrow (Succ_{f,da}^t(\vec{x}) \Leftrightarrow f^{t+1}(\vec{x}))$$

where $Succ_{f,da}^t(\vec{x})$ is a FOL formula at time $t$. One can easily derive the effects of actions by the above axioms.

For example, the precondition of the deterministic action *MvWithObj*($B, ?l_1, ?l_2$) is $At(B, ?l_1) \wedge \neg At(B, ?l_2)$, and its effect is $\neg At(B, ?l_1) \wedge At(B, ?l_2) \wedge (\forall ?o\ In(?o) \Rightarrow At(?o, ?l_2))$.

The grounding of a deterministic action is defined as the grounding of all the fluent atoms appearing in the precondition and effect logical formulas $Precond_{da}^t(\vec{x})$ and $Succ_{f,da}^t(\vec{x})$. Therefore, a grounded deterministic action is a transition function $\mathcal{T} : \mathcal{S} \times D\mathcal{A} \to \mathcal{S}$. One can see from the example that for grounding the action *MvWithObj*, one needs to permute all the possible combinations of *objects* inside the *briefcase*. This increases the dimensionality of the domain and increases the required number of samples to yield low error. Hereinafter for simplicity, we represent fluents and actions without their arguments whenever it is not necessary to mention the variables or domain objects.

**Definition 3.** [Probability distribution for probabilistic actions] Let $\psi_1 \ldots \psi_k$ be FOL formulas (called *partitions*) that divide the world states into mutually disjoint sets. Then, $PA(da|a,s)$ is defined as a probability distribution over possible deterministic executions $da$ of the probabilistic action $a$ in the state $s$ which satisfies one of the partitions $\psi_i$ ($i \leq k$). More formally, when some state $s$ satisfies partition $\psi_i$ then $PA(da|a,s) = PA_i(da)$, where $PA_i$ is a probability distribution over different deterministic executions $da$ of action $a$ corresponding to the partition $\psi_i$.

$$PA(da|a,s) = \begin{cases} PA_1(da) & s \models \psi_1 \\ PA_2(da) & s \models \psi_2 \\ \ldots \end{cases} \quad (1)$$

We assume that replacing variables in $a(\vec{x})$ with constants does not change $PA(da(\vec{x})|a(\vec{x}),s)$.

For example, for probabilistic action *TakeOut*(?o), deterministic action *TakeOutSucc*(?o), and $s \models \psi_1 = In(?o)$: $PA(da|a,s) = PA_1(TakeOutSucc(?o)) = 0.9$.

In PRAM we assume a prior distribution over world states at time 0. Models like [18] are used to represent probabilities in relational models. A knowledge engineer can define the semantics of the prior distribution by using each of these models. Then, we use the inference algorithm defined in that model's semantics to compute probability of formulas at time 0. Our filtering algorithm does not depend on the way the initial knowledge has been represented. We discuss more about this in Section 3.1.3.

In conclusion, we define a PRAM formally as follows:

**Definition 4.** A PRAM is a tuple $(\mathcal{L}, AX, PA, P_0)$ as:

- Language $\mathcal{L} = (F, C, V, \mathcal{A}, D\mathcal{A})$ representing the language of PRAM (Definition 1)
- A set of deterministic action axioms *AX* (Definition 2)
- A probability distribution *PA* for each probabilistic action (Definition 3)
- A prior distribution $P^0$ over initial world states

Based on the aforementioned dynamic system, PRAM, we define our stochastic filtering problem as computing probability of a query given a sequence of probabilistic actions and observations. The query is defined as a FOL formula $\varphi^T$ at the final time step, $T$. Note that throughout the paper superscripts represent time. Each $a^t$ in the probabilistic action sequence $\langle a^1, \ldots, a^T \rangle$ represents the probabilistic action that has been executed at time $t$.

The observations $\langle o^0, \ldots, o^T \rangle$ are given asynchronously in time without prediction of what we will observe (thus,

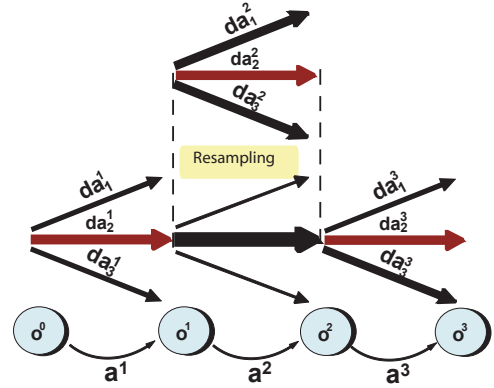

Figure 1: Sampling the FO particle $\langle da_2^1, da_2^2, da_2^3 \rangle$ (straight arrows) given probabilistic sequence $\langle a^1, a^2, a^3 \rangle$ (curved arrows) and observations $\langle o^0, o^1, o^2, o^3 \rangle$. Each deterministic action $da^t$ is sampled from distribution $P(da^t|a^t, da^{1:t-1}, o^{0:t})$. The thickness of the arrows represent the weight of the current FO particle.

this is different from HMMs [16], where a sensor model is given). Each observation $o^t$ is represented with a FOL formula over fluents. When $o^t$ is observed at time $t$, the FOL formula $o^t$ is true about the state of the world at time $t$. We assume that a deterministic execution $da$ of the probabilistic action $a^t$ in the sequence does not depend on the future observations.

## 3 Filtering Algorithm

In this section we present our filtering algorithm for computing probability of a query $\varphi^T$ given a sequence of probabilistic actions $a^{1:T} = \langle a^1, \ldots, a^T \rangle$ and observations $o^{0:T} = \langle o^0, \ldots, o^T \rangle$ in a PRAM. The algorithm approximates this probability by generating samples among possible deterministic executions of the given probabilistic action sequence. Then, it places those samples instead of the enumeration of deterministic executions and marginalizes over those samples. The following equation shows the exact computation.

$$P(\varphi^T|a^{1:T}, o^{0:T}) = \quad (2)$$
$$\sum_i P(\varphi^T|\vec{DA}_i, o^{0:T}) P(\vec{DA}_i|a^{1:T}, o^{0:T})$$

where $\vec{DA}_i$ is a possible execution of the sequence $a^{1:T}$.

The first step of our approximate algorithm is generating $N$ samples (called *FO particles*) from all the possible deterministic executions of the given probabilistic sequence. Two different sampling algorithms are introduced in Section 3.2. The algorithms (illustrated in Figure 1) generate a weighted FO particle $\vec{DA}_i$ with weight $w_i$ given the sequence $a^{1:T}$ and the observations $o^{0:T}$ from the probability distribution $P(\vec{DA}_i|a^{1:T}, o^{0:T})$.

The next step of the algorithm is computing $P(\varphi^T|\vec{DA}_i, o^{0:T})$ for each FO particle $\vec{DA}_i$. It does

**PROCEDURE** *FOFA(SampleAlg, $\varphi^T$, $a^{1:T}$, $o^{0:T}$)*
**Input**: Sampling algorithm *SampleAlg*, probabilistic sequence $a^{1:T}$, observations $o^{0:T}$, and query $\varphi^T$
**Output**: $\tilde{P}(\varphi^T | a^{1:T}, o^{0:T})$
 1. $(\vec{DA}_{1:N}, curF_{1:N}) \leftarrow SampleAlg(a^{1:T}, o^{0:T})$
 2. for each $\vec{DA}_i$ compute $PFOF(\varphi^T, \vec{DA}_i, curF_i)$
 3. **return** $\tilde{P}(\varphi^T | a^{1:T}, o^{0:T})$ using Equation (6).

Figure 2: FOFA: First Order Filtering Algorithm for approximating $P(\varphi^t | a^{1:t}, o^{0:t})$ given a sampling algorithm *SampleAlg* (either *S/R-Actions* (Figure 5) or *S-Actions* (Figure 6)). *Italic fonts* is used to denote subroutines.

so (Section 3.1) by updating the current state of the system and then computing the posterior probability conditioning on the updated current state.

Finally, the algorithm uses generated samples in place of $\vec{DA}_i$ in Equation (2) and computes $\tilde{P}_N(\varphi^T | a^{1:T}, o^{0:T})$ as an approximation for the posterior probability of the query $\varphi^T$ given the sequence $a^{1:T}$ and the observations $o^{0:T}$ by using the Monte Carlo integration [5]:

$$\tilde{P}_N(\varphi^T | a^{1:T}, o^{0:T}) = \sum_i w_i P(\varphi^T | \vec{DA}_i, o^{0:T}) \quad (3)$$

Details of each step of our first order filtering algorithm (*FOFA*, Figure 2) are explained next. We first present the step of computing $P(\varphi^T | \vec{DA}_i, o^{0:T})$ because it is used as a subroutine in the sampling algorithms.

### 3.1 Probability of a FOL Formula at time $t$

In this section we show how to compute the probability $P(\varphi^t | \vec{DA}, o^{0:t})$ of the formula $\varphi^t$ given a FO particle $\vec{DA}$ and observations $o^{0:t}$. Executing the FO particle $\vec{DA}$ with observations $o^{0:t}$ updates the *current state* of the system. The current state is derived by applying a FO *progression* subroutine (described below) at each time step. Afterwards, procedure *PFOF* (Figure 3) computes the probability of the query given the current state of the system for each FO particle. Its first step applies a FO *regression* subroutine to the query and the current state formula and as output returns FOL formulas at time 0. This can be done since the actions are deterministic. The algorithm's second step computes the prior probability of the regression of the query conditioned on the current state formula regressed by the FO particle; Recall that a FO particle is a sampled sequence of deterministic actions.

#### 3.1.1 Progress Current State Formula

We define the current state formula as a FOL formula representing the set of states that are true after executing a sequence of deterministic actions and receiving observations. In this section we present algorithm *Progress* that updates the current state formula given a deterministic action and an observation. In general, progressing a FOL for-

**PROCEDURE** *PFOF($\varphi^t$, $\vec{DA}$, curF)*
**Input**: FO particle $\vec{DA}$ and the current state formula *curF*
**Output**: $P(\varphi^t | \vec{DA}, o^{0:t})$
 1. $\varphi^0 \leftarrow RegSeq(\varphi^t, \vec{DA})$
 2. $curF^0 \leftarrow RegSeq(curF, \vec{DA})$
 3. $p_0 \leftarrow Prior\text{-}FOF(\varphi^0 | curF^0)$
 4. **return** $P(\varphi^t | \vec{DA}, o^{0:t}) \leftarrow p_0$

**PROCEDURE** *Prior-FOF($\varphi^0$, $M_{MLN}$)*
**Input**: Formula $\varphi^0$, Markov network $M_{MLN}$ of the prior MLN
**Output**: $P(\varphi^0)$
 1. $\bigwedge_i C_i \leftarrow ConvertToCNF(\varphi^0)$
 2. Define indicator functions as Equation (4)
 3. $M \leftarrow ConstructNetwork(M_{MLN}, \text{fluents of } \varphi^0)$
 4. **return** $P(\varphi^0)$ by performing inference on $M$

Figure 3: *PFOF*: Compute probability $P(\varphi^t | \vec{DA}, o^{0:t})$ of a FOL formula $\varphi^t$ given a FO particle $\vec{DA}$ and the current state formula.

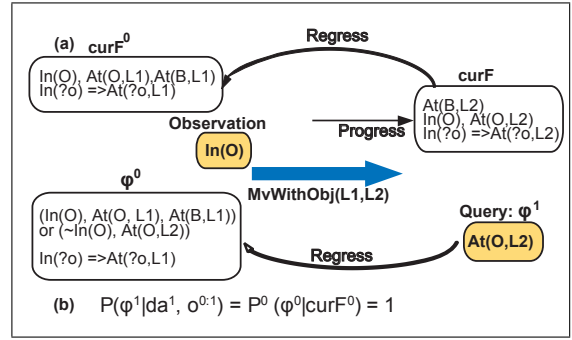

Figure 4: (a) Regressing query $\varphi^1$ and *curF* by *MvWithObj*. Note, $L1$, $L2$, and $O$ are constants, and there are only two possible locations $L1$ and $L2$, i.e. $At(O, L2) \equiv \neg At(O, L1)$. (b) Computing $P(\varphi^2 | \vec{DA}, o^{0:1})$ using Lemma 1.

mula $\delta$ with a deterministic action $da$, $Progress(\delta, da)$, results in a set of FOL formulas $\Delta(p_{1:n})$ where $p_1, \ldots, p_n$ are atomic subformulas of $\Delta$ and $\delta \wedge Precond_{da} \models \Delta(Succ_{p_1, da}, \ldots, Succ_{p_n, da})$ (see [19] for more details). Also, filtering a FOL formula $\delta$ with an observation $o$ is the FOL formula $\delta \wedge o$.

Overall, generating all $\Delta$s is impossible because there are infinitely many such $\Delta$s. In this paper we assume that progression of the current state formula *curF* with a deterministic action $da$, $Progress(curF, da, o)$, is representable with a FOL formula. Hence, it is equal to $\bigvee_i \Delta_i(p_{1:n}) \wedge o$. Furthermore, [19; 14] provide some special conditions that the progression algorithm is polynomial and the representation of progressing a formula is compact. For example, progressing $In(O)$ through deterministic action $MvWithObj(L1, L2)$ results in the formula $At(B, L2) \wedge In(O) \wedge At(O, L2) \wedge (\forall ?o \, In(?o) \Rightarrow At(?o, L2))$.

#### 3.1.2 Regressing a FOL Formula

Procedure *RegSeq* takes a FOL formula $\varphi^t$ and a FO particle $\vec{DA}$ and returns as output another FOL formula $\varphi^0$. $\varphi^0$ represents the set of possible initial states, given that the fi-nal state satisfies $\varphi^t$, and the FO particle $\vec{DA}$ occurs. Thus, every state that satisfies $\varphi^0$ leads to a state satisfying $\varphi^t$ after $\vec{DA}$ occurs.

For a deterministic action $da^t$ and a FOL formula $\varphi^t$, the regression $Regress(\varphi^t, da^t)$ of $\varphi^t$ through $da^t$ is a FO formula $\varphi^{t-1}$ such that state $s^{t-1}$ satisfies $\varphi^{t-1}$ iff the result of the transition function $\mathcal{T}(s^{t-1}, da^t)$ satisfies $\varphi^t$. The computation of the regression $RegSeq(\varphi^t, \vec{DA})$ of $\varphi^t$ through the FO particle $\vec{DA}$ is done recursively.

$$RegSeq(\varphi^t, \langle da^1, ..., da^t \rangle) = \\ RegSeq(Regress(\varphi^t, da^t), \langle da^1, ..., da^{t-1} \rangle).$$

Regression of the current state formula *curF* is defined similar to the regression of formula $\varphi^t$ since the current state is also represented with a fist order formula.

The algorithm for regression $Regress(\varphi^t, da)$ of formula $\varphi^t$ with deterministic action $da$ works as follows (see [17]). Suppose that $\varphi^t$ is an atomic fluent $f^t(\vec{x})$ with the successor state axiom $Succ_{f,da}^{t-1}(\vec{x})$. Then, we derive the regression of $\varphi^t$ by replacing $f^t(\vec{x})$ with $Succ_{f,da}^{t-1}(\vec{x})$. The regression of non-atomic formulas are derived inductively as follows:

- $Regress(\neg \varphi^t) = \neg Regress(\varphi^t)$
- $Regress(\varphi_1^t \wedge \varphi_2^t) = Regress(\varphi^t) \wedge Regress(\varphi_2^t)$
- $Regress((\exists \nu)\varphi^t) = \exists \nu Regress(\varphi^t)$

The efficiency of the regression yields from the fact that we regress non-grounded fluents $f(\vec{x})$. Besides, one can employ some approximations at each step (like removing unnecessary clauses) to maintain the compactness of the regressed formulas.

We now summarize and show how to compute the probability distribution $P(\varphi^t | \vec{DA}, o^{0:t})$ by applying progression and regression. The algorithm first computes *curF* by progressing through the FO particle and observations. Then, it computes $\varphi^0$ and $curF^0$ by regressing $\varphi^t$ and *curF*. Finally, it computes $P^0(\varphi^0 | curF^0)$ as the evaluation of $P(\varphi^t | \vec{DA}, curF)$ as shown by the following lemma.

**Lemma 1.** Let $\varphi^t$ be the query and $o^{0:t}$ the observations. If *curF* is the current state formula, $\varphi^0 = RegSeq(\varphi^t, \vec{DA})$, and $curF^0 = RegSeq(curF, \vec{DA})$, then $P(\varphi^t | \vec{DA}, o^{0:t}) = P^0(\varphi^0 | curF^0)$.

*Proof.* Probability of a formula is a marginalization over states: $P(\varphi^t, o^{0:t} | \vec{DA}) = \sum_s P(\varphi^t, curF | s, \vec{DA}) P(s | \vec{DA})$. For $s$, $P(\varphi^t, curF | s, \vec{DA}) = 1$ if $s \models \varphi^0 \wedge curF^0$, o.w. it is 0. The reason is that executing the deterministic sequence $\vec{DA}$ in state $s$ results at time $t$ that models $\varphi^t$ and is consistent with observations $o^{0:t}$ iff $s \models \varphi^0 \wedge curF^0$. Therefore, $P(\varphi^t, o^{0:t} | \vec{DA}) = \sum_{s \models \varphi^0, curF^0} P(s) = P^0(\varphi^0, curF^0)$. The same computations exist for $P(o^{0:t} | \vec{DA})$. Therefore,

$$P(\varphi^t | \vec{DA}, o^{0:t}) = \frac{P(\varphi^t, o^{0:t} | \vec{DA})}{P(o^{0:t} | \vec{DA})} = \frac{P^0(\varphi^0, curF^0)}{P^0(curF^0)}. \quad \blacksquare$$

Figure 4 shows an example of computing $P(\varphi^t | \vec{DA}, o^{0:t})$ using procedure *PFOF*. The next section shows how to compute $P^0(\varphi^0 | curF^0)$ using the prior $P^0$. Note that $\varphi^0$ and $curF^0$ are FOL formulas over fluents at time 0.

### 3.1.3 Prior Probability of a FOL Formula

In this section we describe procedure *Prior-FOF* to compute the probability of a FOL formula at time 0. We assume that the prior distribution $P^0$ over world states of a PRAM with language $\mathcal{L} = (F, C, V, \mathcal{A}, D\mathcal{A})$ is represented by a Markov Logic Network (MLN) [18]. We choose to represent our prior probabilistic logic with an MLN (called *prior MLN*) because it keeps the expressive power of FOL for a fixed domain.

Our prior MLN consists of a set of weighted FOL formulas over the fluents in language $\mathcal{L}$ of the PRAM. The semantics of the MLN is that of a Markov network $M_{MLN}$ [9] whose cliques correspond to groundings of the formulas given the universe of objects $C \in \mathcal{L}$. The potential $\Phi$ of a clique $Cl$ is defined as the exponential of the weight of the corresponding formula in case the grounding is true.

We introduce *Prior-FOF* (Figure 3) to compute the probability of a FOL formula $\varphi^0$ by slightly changing the inference algorithm expressed for MLNs. Procedure *Prior-FOF* first converts $\varphi^0$ to a clausal form; Note that like MLNs existentially quantified formulas are replaced by disjunction of their groundings. Then, *Prior-FOF* assigns an indicator function, $\mathcal{I}_{g(C_i)}$, to every grounding $g$ of a clause $C_i$.

$$\mathcal{I}_{g(C_i)}(\vec{f}_{g(C_i)}) = \begin{cases} 1 & \vec{f}_{g(C_i)} \models g(C_i) \\ 0 & \text{otherwise} \end{cases} \quad (4)$$

The next step is to construct a Markov network $M$ as the minimal subset of the original network $M_{MLN}$ required to compute $P^0(\varphi^0)$. The nodes in $M$ consist of all the ground fluents $f_{g(C_i)}$ in clauses $C_i$ of $\varphi^0$ and all the nodes in the original Markov network $M_{MLN}$ with a path to the fluent $f_{g(C_i)}$. The final step is to perform inference on this network. It can be computed exactly by performing variable elimination (e.g., [9]) in $P^0(\varphi^0) \propto \sum_{f \in M} \prod_{\vec{f}_j \in Cl_j, \vec{f}_i \in C_i} \Phi_j(\vec{f}_j) \mathcal{I}(\vec{f}_i)$. Also, one can employ any approximate inference algorithm like Gibbs sampling.

We are not restricted to use MLNs as a framework for representing prior probability distribution. We can use any other probabilistic logical frameworks ([3; 12; 15; 7; 20]) provided that the expressed inference algorithm in that framework can compute probability of the query. Note that the query $\varphi^0$ for that framework is derived from the regression of the original query $\varphi^t$ given an FO particle. Also, our algorithms would work for unbounded domains just by using a framework (e.g., [21]) for representing prior distribution over infinite states.

```
PROCEDURE S/R-Actions( a^{1:T}, o^{0:T} )
Input: Probabilistic sequence a^{1:T} and observations o^{0:T}
Output: N FO particles DA_{1:N}
  1. for n ← 1...N : curF_n ← o^0
  2. for t ← 1...T
  3.   for n ← 1...N
  4.     π ← ∑_{ψ_i} PA_i · PFOF(ψ_i^{t-1}, da^{1:t-1})
  5.     P ← ∑_{ψ_i} PA_i · PFOF(ψ_i^{t-1}, da_n^{1:t-1}, curF_n)
  6.     dâ^t ← a sample from probability distribution π
  7.     curF_n ← Progress(curF_n, dâ^t_n, o^t)
  8.     w*_n ← π(dâ^t_n) / P(dâ^t_n)
  9.   ⟨ŵ_1, ..., ŵ_N⟩ ← Normalize(w*_1, ..., w*_N)
 10.   (da^t_{1:N}, w_{1:N}) ← Resample(dâ^t_{1:N}, ŵ_{1:N})
 11. return DA_{1:N} ← ⟨da^1 ... da^T⟩_{1:N}
```

Figure 5: *S/R-Actions*: Sampling/Resampling algorithm for generating $N$ FO particles given a sequence $a^{1:T}$ and observations $o^{0:T}$.

```
PROCEDURE S-Actions( a^{1:T}, o^{0:T} )
Input: Probabilistic sequence a^{1:T} and observations o^{0:T}
Output: N FO particles DA_{1:N}
  1. for n ← 1...N : curF_n ← o^0
  2. for t ← 1...T
  3.   for n ← 1...N
  4.     P ← ∑_{ψ_i} PA_i · PFOF(ψ_i^{t-1}, da^{1:t-1}, curF_n)
  5.     da^t ← a sample from probability distribution P
  6.     curF_n ← Progress(curF_n, da^t, o^t)
  7. return DA_{1:N} ← ⟨da^1 ... da^T⟩_{1:N}
```

Figure 6: *S-Actions*: Direct sampling algorithm for generating $N$ FO particles given a sequence $a^{1:T}$ and observations $o^{0:T}$.

### 3.2 Sampling Algorithms

In this section we describe procedures *S/R-Actions* (Figure 5) and *S-Actions* (Figure 6) which generate $N$ samples (called *FO particles*) given a sequence of probabilistic actions and observations. Each FO particle is a possible deterministic execution of the given probabilistic sequence. Both algorithms incrementally build every FO particle by sampling a deterministic action at a time. Later, we will discuss about the deficiencies of *S/R-Actions* algorithm.

Both *S/R-Actions* and *S/Actions* generate a FO particle $\vec{DA} = \langle da^1, \ldots, da^T \rangle$ given a sequence $a^{1:T}$ and observations $o^{0:T}$ from the distribution $P(\vec{DA}|a^{1:T}, o^{0:T})$. We compute this probability distribution iteratively:

$$P(\vec{DA}|a^{1:T}, o^{0:T})$$
$$= P(da^1|a^1, o^{0:1}) \prod_t P(da^t|a^t, da^{1:t-1}, o^{0:t})$$

The above derivation allows iterative sampling of deterministic actions from the distribution $P(da^t|a^t, da^{1:t-1}, o^{0:t})$. Recall that a FO particle is a sequence of deterministic actions. Thus, at each time the algorithms sample a deterministic action $da^t$ given the probabilistic action $a^t$, the previous deterministic actions $da^{1:t-1}$, and the previous observations $o^{0:t}$; Note that current deterministic action is independent of the future observations. Then, the algorithms update the current state formula *curF* (Section 3.1.1) given the current deterministic action and the observation.

Algorithms *S/R-Actions*, *S-Actions* use different approaches for incremental sampling of each deterministic action. In *S/R-Actions*, we adopt a sequential importance sampling approach. At each time step, *S/R-Actions* samples deterministic actions based on a normalized importance function, assigns weights to the particles, and resamples if the weights have high variance. The importance function $π(da^t|a^t, da^{1:t-1})$ approximates $P(da^t|a^t, da^{1:t-1}, o^{0:t})$ by ignoring the effect of the current state *curF* in the $PRAM = (\mathcal{L}, AX, PA, P^0)$.

$$π(da^t|a^t, da^{1:t-1}) = \sum_i PA_i(da^t) PFOF(ψ_{i,a^t}^{t-1}, da^{1:t-1})$$

where $ψ_{i,a^t}$ is the $i^{th}$ partition of action $a^t$ (Definition 3).

At time step $t$, *S/R-Actions* samples $N$ deterministic actions $da^t_{1:N}$ from the importance function: $π(da^t|a^t, da^{1:t-1})$. It then restructures the $n^{th}$ particle $\vec{DA}_n^{t-1}$ by attaching the deterministic action $da^t_n$ to that particle, i.e. $\vec{DA}_n^t = \langle da_n^1, \ldots, da_n^{t-1}, da_n^t \rangle$. The importance weight $w^*_n$ of the $n^{th}$ particle is derived as: $w^*_n = \frac{P(da^t_n|a^t, da_n^{1:t-1}, o^{0:t})}{π(da^t_n|a^t, da_n^{1:t-1})}$.

Accordingly, the normalized importance weight $w_n$ of the $n^{th}$ particle is: $w_n = \frac{w^*_n}{\sum_{i=1}^N w^*_i}$.

The exact value for $P(da^t|a^t, da^{1:t-1}, o^{0:t})$ is computed using *PFOF* subroutine (Figure 3).

$$P(da^t|a^t, da^{1:t-1}, o^{0:t}) \quad (5)$$
$$= \sum_i PA_i(da^t) \cdot PFOF(ψ_{i,a^t}^{t-1}, da^{1:t-1}, curF)$$

*S/R-Actions* resamples the particles if the variance of the normalized importance weights becomes too high. The basic idea of resampling is to avoid those particles with very low weight and concentrate on particles that have higher normalized weight. An estimation (see [5]) for measuring high variance among the weights is estimating the *effective number* of particles as $\widehat{N_{eff}} = \frac{1}{\sum_{i=1}^N w_i}$. If $\widehat{N_{eff}}$ is smaller than a threshold then procedure *Resample* generates a new deterministic action $da^t$ from the probability distribution over the normalized weights $w_{1:N}$ of the particles. It then assigns equal weights, $\frac{1}{N}$, to all the particles.

The second sampling algorithm, *S-Actions* (Figure 6), is derived by simplifying the above sampling algorithm. This algorithm at each time step generates samples among executable deterministic actions. It samples every deterministic action in a FO particle from the exact computation for $P(da^t|a^t, da^{1:t-1}, o^{0:t-1})$ (Equation 5) instead of sampling from the importance function and assigning weights. Therefore, all the weights are equal to $\frac{1}{N}$.

*S/R-Actions*, samples deterministic actions even when they are not executable and assigns weight zero to those particles. Therefore, it decreases the effective number of samples. Resampling step does not help that much because it resamples the latest deterministic action in the FO particle. However, resampling earlier deterministic actions may result in a more effective set of FO particles. Note that this new type of resampling is not tractable. Therefore, it makes more sense to maintain the current state formula *curF* as in *S-Actions* and just sample executable deterministic actions (as in *S-Actions*) unless there exists a better resampling procedure. In the empirical results we examine the effects of using *S/R-Actions* and *S-Actions* sampling algorithms in the *FOFA* filtering algorithm (Figure 2).

### 3.3 Correctness, Complexity, and Accuracy

The following theorem shows how *FOFA* with *S-Actions* and *S/R-Actions* compute the approximate posterior distribution $\tilde{P}_N(\varphi^T|a^{1:T}, o^{0:T})$.

**Theorem 1.** Let $\varphi^T$ be the query, $a^{1:T}$ be the given probabilistic sequence, $\vec{DA}_i$ be the FO particles, and $o^{0:T}$ be the observations. If $curF_i$ is the current formula given the $i^{th}$ FO particle, $\varphi_i^0 = RegSeq(\varphi^T, \vec{DA}_i)$, and $curF_i^0 = RegSeq(curF_i, \vec{DA}_i)$, then

$$\tilde{P}_N(\varphi^T|a^{1:T}, o^{0:T}) = \sum_i w_i P^0(\varphi_i^0|curF_i^0) \quad (6)$$

where $w_i = \frac{1}{N}$ for *S-Actions*. Besides, for *S-Actions*:

$$\tilde{P}_N(\varphi^T|a^{1:T}, o^{0:T}) \to_{N\to\infty} P(\varphi^T|a^{1:T}, o^{0:T}) \quad (7)$$

The proof follows from using MC integration in Equation 2 and using Lemma 1.

The running time $R_{FOFA}$ of our filtering algorithm (Figure 2) is $O(N \cdot T \cdot (R_{RegSeq} + R_{Progress} + R_{Prior-FOF}))$, where $N$ is the number of samples and $T$ is the length of the given sequence of probabilistic actions. Efficiency of *FOFA* results from efficiency of the underlying algorithms for *RegSeq*, *Progress*, and *Prior-FOF*.

We evaluate the accuracy of our sampling algorithm *FOFA* by computing expected KL-distance [1] as the expected value of all the KL-distances between the exact distribution $P$ and the approximation $\tilde{P}$ derived by *FOFA*. Our algorithm, *FOFA*, has higher accuracy than SMC for a fixed number of samples. The intuition is that each FO particle generated by *FOFA* covers many particles generated by SMC.

**Theorem 2.** If *FOFA*(*S-Actions*) and SMC approximate posterior distribution $P(\varphi^T|a^{1:T}, o^{0:T})$ with $N$ samples. Then, Expected-KL$_{FOFA(S\text{-}Actions)} \leq$ Expected-KL$_{SMC}$.

[1] $KL(P, \tilde{P}) = \sum_x P_x log(P_x/\tilde{P}_x), KL(P, \tilde{P}) = 0$ if $P = \tilde{P}$

The proof is similar to the proof of Theorem 3.3 in [8]. Intuitively, we define a mapping $f$ to map each set of FO particles of *S-Actions* to sets of particles of SMC. The mapping $f$ is defined such that it covers all the possible sets of particles of SMC, and for two separate sets of particles $z_i \neq z_j$, $f(z_i) \cap f(z_j) = \emptyset$, and $Pr_{FOFA}(z) = Pr_{SMC}(f(z))$. Furthermore, we prove that $\forall y \in f(z), KL(P, \tilde{P}_1^z) \leq KL(P, \tilde{P}_2^y)$ where $\tilde{P}_1$ and $\tilde{P}_2$ are approximations returned by *FOFA* and SMC, respectively.

## 4 Empirical Results

We implemented our algorithm *FOFA* (Figure 2) with both *S/R-Actions* (Figure 5) and *S-Actions* (Figure 6). Our algorithms take advantage of a different structure than that available in DBNs. Hence, we focused on planning-type domains: *briefcase* and *depots* taken from International Planning Competition at AIPS-98 and AIPS-02. [2] We randomly assigned deterministic executions and a probability distribution over them for each action. For example, for action *PutIn* we considered two executions *PutInSucc* and *PutInFail* with probabilities 0.9 and 0.1. Note that DBN representations (transitions between states) for the above frameworks are not compact because the independence assumptions among the state variables are not known.

We compared the accuracy of our *FOFA*(*S/R-Actions*) and *FOFA*(*S-Actions*) with SCAI [8] and SMC algorithms. Note that we grounded the domains for running SCAI and SMC. We ran sampling algorithms (50 times) for a fixed number of samples and computed the KL-distance between their approximation and the exact posterior. We calculated the average over these derived KL-distances to approximate the expected KL-distance. SCAI assumes that deterministic actions are always executable. We include this assumption in the *depots* domain in which we compared the results with SCAI. In the *briefcase* domain, we just compared our algorithms with the SMC techniques.

Figures 7 and 8 show the expected KL-distances (in logarithmic scale) vs. number of samples in the *depots* and *briefcase*, respectively. As we expected, the average KL-distance for *FOFA*(*S-Actions*) is always the lowest. We expect to get more improvement when there are more objects in the domain (here, we just use 5 constants to be able to compute the exact distribution). For the *briefcase* domain (Figure 8) SMC has higher accuracy than *FOFA*(*S/R-Actions*) for 1000 samples. The reason is that the posterior distribution converges to the stationary distribution in this domain (4 constants, 256 states). Even grounded domain is not too big, but for cases involving longer sequences and more states we did not have the exact posterior to compare with since the implementation for the exact algorithm crashes.

[2] Also available from: ftp://ftp.cs.yale.edu/pub/mcdermott/domains/.

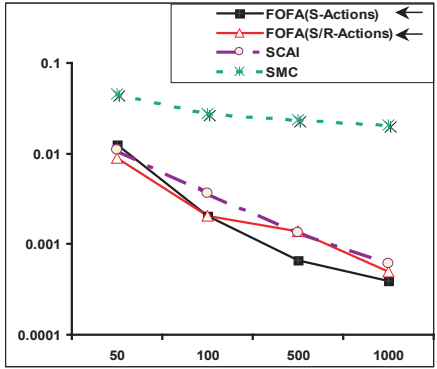

Figure 7: Expected KL-distance (in **logarithmic scale**) of our algorithms, *FOFA*(*S-Actions*) and *FOFA*(*S/R-Actions*), SCAI, and SMC with the exact distribution vs. number of samples for *depots*.

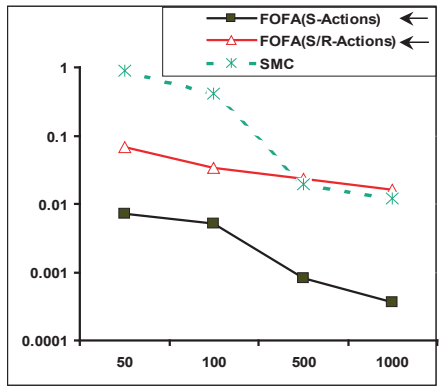

Figure 8: Expected KL-distance (in **logarithmic scale**) of our algorithms, *FOFA*(*S-Actions*) and *FOFA*(*S/R-Actions*), and SMC with the exact distribution vs. number of samples for *briefcase*.

## 5 Conclusions and Future Work

In this paper we presented a sampling algorithm to compute the posterior probability of a query given a sequence of actions and observations in a FO dynamic system. Our algorithm takes advantage of a compact representation and achieves higher accuracy than SMC sampling and earlier propositional sampling techniques.

There are several directions that we can continue this work: (1) Apply sampling in FO Markov Decision Processes(MDP)s [2](2) Learn the transition model in PRAM. (3) Apply the algorithm in text understanding. (4) Generalize the representation to continuous domains (e.g., by discretizing the real value variables or by combining with Rao-Blackwellised Particle Filtering [6]).

**Acknowledgements**

We would like to thank the anonymous reviewers for their helpful comments. This work was supported by DARPA SRI 27-001253 (PLATO project), NSF CAREER 05-46663, and UIUC/NCSA AESIS 251024 grants.